\def\year{2022}\relax
\documentclass[letterpaper]{article} 
\usepackage{subcaption}
\usepackage{aaai22}  
\usepackage{times}  
\usepackage{helvet}  
\usepackage{courier}  
\usepackage[hyphens]{url}  
\usepackage{graphicx} 
\usepackage{booktabs}
\usepackage{natbib}  
\usepackage{caption} 
\DeclareCaptionStyle{ruled}{labelfont=normalfont,labelsep=colon,strut=off} 
\usepackage{algorithm}
\usepackage{algorithmic}

\usepackage{mathrsfs}
\usepackage{amsthm}

\newtheorem{definition}{Definition}
\usepackage{amsmath}
\usepackage{stmaryrd}
\newcommand\numberthis{\addtocounter{equation}{1}\tag{\theequation}}

\usepackage{newfloat}
\usepackage{listings}
\floatstyle{ruled}
\newfloat{listing}{tb}{lst}{}
\floatname{listing}{Listing}

\title{Blackbox Postprocessing for Multiclass Fairness}
\author{
    Preston Putzel\textsuperscript{\rm 1}\equalcontrib,
    Scott Lee\textsuperscript{\rm 2}\equalcontrib
}
\affiliations{
    \textsuperscript{\rm 1}Department of Computer Science, University of California, Irvine, CA, 92697, USA\\
    \textsuperscript{\rm 2}Centers for Disease Control and Prevention, 1600 Clifton Rd., Atlanta, GA, USA
    
}

\begin{document}

\maketitle

\begin{abstract}
Applying standard machine learning approaches for classification can produce unequal results across different demographic groups. When then used in real-world settings, these inequities can have negative societal impacts. This has motivated the development of various approaches to fair classification with machine learning models in recent years. In this paper, we consider the problem of modifying the predictions of a blackbox machine learning classifier in order to achieve fairness in a multiclass setting. To accomplish this, we extend the 'post-processing' approach in \citet{hardt2016equality}, which focuses on fairness for binary classification, to the setting of fair multiclass classification. We explore when our approach produces both fair and accurate predictions through systematic synthetic experiments and also evaluate discrimination-fairness tradeoffs on several publicly available real-world application datasets. We find that overall, our approach produces minor drops in accuracy and enforces fairness when the number of individuals in the dataset is high relative to the number of classes and protected groups.
\end{abstract}



\section{Introduction}

As machine learning begins moving into sensitive predictions tasks, it becomes critical to ensure the fair performance of prediction models. Naively trained machine learning systems can replicate biases present in their training data, resulting in unfair outcomes that can accentuate societal inequities. For example, machine learning systems have been discovered to be unfair in predicting time to criminal recidivism \citep{dieterich2016compas}, ranking applications to nursing school \citep{romano2020achieving}, and recognizing faces \citep{buolamwini2018gender}. Most prior work in this area has focused on ensuring fairness for binary outcomes. However, there are many important real-world applications with multiclass outcomes instead. For example, a self-driving car will need to be able to distinguish clearly between humans, non-human animals (such as dogs), and non-sentient objects while nonetheless maintaining fair performance for both wheelchair users and non-wheelchair users. 
Most work has also been done with the assumption that model parameters are accessible to the algorithm, but there is increasing availability of powerful blackbox models whose internal parameters can be either inaccessible or too costly to train. In this paper, we address the case where outcomes are multiclass and the user has received a pre-trained blackbox model. The main contributions of our work are as follows:
\begin{itemize}
    \item We show how to extend \citet{hardt2016equality} to multiclass outcomes.
    \item We demonstrate in what data regimes multiclass postprocessing is likely to produce fair, useful, and accurate results via a set of rigorous synthetic experiments.
    \item We demonstrate the results of our post-processing algorithm on publicly available real-world applications.
\end{itemize}

\paragraph{Code and Dataset Availability} All of the code used to produce our experimental results as well as the synthetic and real-world datasets can be found on our github page\footnote{https://github.com/scotthlee/fairness/tree/aaai}. 
\section{Technical Approach}
As in \citet{hardt2016equality}, we consider the problem of enforcing fairness on a blackbox classifier without changing its internal parameters. This means that our approach only has access to the predicted labels $\hat{y_i}$ from the blackbox classifier, the true labels $y_i$, and the protected attributes $a_i$ for $i\in\{1, ..., N\}$ where $N$ is the number of individuals. The goal of our approach is to produce a new set of updated and fair 'adjusted' predictions $y^{\text{\text{adj}}}_i$ that satisfy a desired fairness criterion.  For each of $\hat{y_i}$, $y_i$, and $a_i$, we define corresponding random variables $\hat{Y}$, $Y$, $A$. Then, following \citet{hardt2016equality} we define the random variable for the adjusted predictions $Y^{\text{\text{adj}}}$ to be a randomized function of $\hat{Y}$ and $A$. 
We extend the approach in \citet{hardt2016equality} by allowing multiclass outcomes, such that the sample spaces of $\hat{Y}$, $Y$, and $Y^{\text{\text{adj}}}$ are a collection of discrete and mutually exclusive outcomes $\mathcal{C} = \{1, 2, .... , |C|\}$. We in principle allow the sample space of the protected group $A$, $\mathscr{A}$, to contain any number of discrete values as well: $\mathscr{A} = \{1, 2, ..., |\mathscr{A}|\}$.

\paragraph{Linear Program} Our approach involves the construction of a linear program over the conditional probabilities of the adjusted predictor $Pr(Y^{\text{\text{adj}}}=y^{\text{\text{adj}}}|\hat{Y}=\hat{y}, A=a)$ such that a desired fairness criterion is satisfied by those probabilities. In order to construct the linear program, both the loss and fairness criteria must be linear in terms of the protected attribute conditional probability matrices ${\mathbf{P^{a}} =Pr(Y^{\text{adj}}|\hat{Y}, A=a)}$ which have dimensions $|C|\times |C|$. 

\paragraph{Types of Objective Functions} We consider objective functions which are linear in the group conditional adjusted probabilities $\mathbf{P^{a}}$. More specifically we consider minimizing expected losses of the form:
\begin{align*}
     &E[l(y^{\text{adj}}, y)] = \\  &\sum_{a\in\mathscr{A}}\sum_{i=1}^{|C|} \sum_{j\neq i} Pr(Y^{\text{adj}}=i, Y=j, A=a)l(i, j, a) \\
    &= \sum_{a\in\mathscr{A}}\sum_{i=1}^{|C|} \sum_{j\neq i} W^{a}_{ij}\;\;Pr(A=a, Y=j)\;l(i, j, a) 
\end{align*}
 where $W^a_{ij}=Pr(Y^{\text{adj}}=i|Y=j, A=a)$ are the protected attribute conditional confusion matrices. 
 Under the independence assumption $Y^{\text{adj}} \perp Y | A, \hat{Y}$, we can write $\mathbf{W^a}=\mathbf{P^a} \mathbf{Z^a}$ where $\mathbf{Z^a} = Pr(\hat{Y}|Y, A=a)$, the class conditional confusion matrices of the original blackbox classifier's predictions. The matrices $\mathbf{Z^a}$ are estimated empirically from the training data ($y_i$, and $a_i$) and blackbox predictions of the model ($\hat{y_i}$). Therefore, this formulation of the objective function remains linear in the protected attribute conditional probability matrices, $\mathbf{P^a}$, as is necessary for the linear program. This definition is similar to \citet{hardt2016equality} except we let the loss $l(i, j, a)$ also be a function of protected attributes instead of just the true and adjusted labels, which allows controlling the strictness of penalties for errors made for specific protected groups and classes. The most straightforward version of this loss is letting $l(y^{\text{adj}}, y, a)$ be the zero-one loss (ignoring the protected attributes) which results in minimizing the sum of the joint probabilities of mismatch between $Y^{\text{adj}}$ and $Y$. We refer to this approach as \textit{unweighted} loss. Another approach is to set $l(y^{\text{adj}}, y, a)$ equal to one over the joint probabilities of the true label and protected attribute $1/Pr(Y=y, A=a)$ (estimated empirically), which we refer to as \textit{weighted} loss. Intuitively, this option reweights the loss to give rarer protected groups and label combinations equal importance to the optimization which could improve fairness when very low membership minority protected groups exist in the dataset.  This option for the objective function can be equivalently minimized by maximizing the diagonals (true detection rates) of the group conditional confusion matrices $\mathbf{W^a}$.

\paragraph{Types of Fairness}  We consider several versions of multiclass fairness criteria, all of which can be written as a collection of $|\mathscr{A}| - 1$ pairwise equalities setting a fairness criterion of interest equal across all groups. Moreover, each of the terms in these equalities can be written as some $|C|\times|C|$ matrix $M^a$ times the adjusted probability matrix $\mathbf{P^a}$, and therefore are linear in the adjusted probabilities as needed for the linear program (see appendix A for the exact form $M^a$ takes for the different fairness criteria). 

The first definition involves requiring strictly equal performance across protected groups.
\begin{definition}[Term-by-Term Multiclass Equality of Odds]
A multiclass predictor satisfies term-by-term equality of odds if the protected group conditional confusion matrices $\mathbf{W^a}$ are equal across all protected groups:
\begin{equation}
\label{def.strict}
    \mathbf{W^{1}}= \mathbf{W^{2}}=\dots=\mathbf{W^{|\mathscr{A}|}}
\end{equation}
where $\mathbf{W^{a}}=Pr(Y^{\text{adj}}|Y, A=a)$.
\end{definition}
This is a straightforward extension to the multiclass case of equality of odds defined in \citet{hardt2016equality}. Notice that since this definition requires equality of each off-diagonal term of $\mathbf{W^a}$ across all groups, it enforces that not only are errors made at the same overall rate across groups, but also that the rate of specific types of errors are equal.
For some practical applications, term-by-term equality of odds is important, such as predicting criminal recidivism times binned into three years, two years, one year, and "never recommits". In this case, making the error of predicting 3 years until recidivism when the actual time is 1 year is much worse than predicting 3 years when the actual time is 2. Therefore, it is critical for fairness in this application that the rates of specific types of errors are strictly equal across groups. 

Instead of requiring strict equality of off-diagonal terms of $\mathbf{W^a}$ we can instead enforce equality across the classwise overall false detection rates $FDR$, which leads to the next fairness definition:

\begin{definition}[Classwise Multiclass Equality of Odds]
A multiclass predictor satisfies classwise multiclass equality of odds if the diagonals of the protected group conditional confusion matrices and the protected attribute conditional vector of false detection rates are equal across all protected groups:
\begin{equation*}
\label{def.relaxed}
\begin{array}{l}
    diag(\mathbf{W^{1}})= diag(\mathbf{W^{2}})=\dots=diag(\mathbf{W^{|\mathscr{A}|}})\\
    \mathbf{FDR^{1}} = \mathbf{FDR^{2}} = \dots = \mathbf{FDR^{|\mathscr{A}|}}\numberthis
\end{array}
\end{equation*}
where $\mathbf{FDR^{a}} = Pr(Y^{\text{adj}}|Y^{\text{adj}}\neq Y, A=a)$.
\end{definition}
This version of fairness can 'trade' better performance  for a specific protected group on one off diagonal term in $\mathbf{W^a}$ (i.e. lower error probability for that term) for poorer performance of the same group on a different off diagonal term (i.e. higher error probability for another term). Individually each class label has it's true detection rate, and overall false detection rate set equal across groups. Thus, this type of fairness is 'classwise'.

For some problems it is sufficient to maintain fair true detection rates across classes and allow false detection rates to differ across groups. This is even less restrictive than Definition \ref{def.relaxed}. This may be desirable when, for example, deciding whether an accepted college application should be accepted into a honors program, accepted with scholarship, or regularly accepted. Since all the outcomes are positive, unfairness across false detection rates may not be critical, as long as the true detection rates are fair across groups. This motivates the following fairness criteria:
\begin{definition}[Multiclass Equality of Opportunity]
A multiclass predictor satisfies equality of opportunity if the diagonals of the protected group conditional confusion matrices $\mathbf{W^a}$ are equal across all groups:
\begin{equation}
    diag(\mathbf{W^{1}})= diag(\mathbf{W^{2}})=\dots=diag(\mathbf{W^{|\mathscr{A}|}})
\end{equation}
where $\mathbf{W^{a}}=Pr(Y^{\text{adj}}|Y, A=a)$.
\end{definition}

A common and even more relaxed version of fairness called demographic parity only requires the rate of class predictions across different groups to be equal \citep{calders2009building}.
\begin{definition}[Multiclass Demographic Parity]
A multiclass predictor satisfies demographic parity if the protected group conditional class probabilities are equal across groups:
\begin{equation}
\begin{array}{l}
Pr(Y^{\text{adj}}|A=1)=\\
Pr(Y^{\text{adj}}|A=2) =\dots =Pr(Y^{\text{adj}}|A=|\mathscr{A}|)
\end{array}
\end{equation}
\end{definition}
Enforcing this version of fairness for certain datasets may produce effectively unfair outcomes \citep{dwork2012fairness}. However, in synthetically produced data, this definition has been shown to reduce the reputation of disadvantaged protected groups when repeatedly applied over a long period of time to sensitive decision-making tasks such as hiring \citep{hu2018short}.

Note that while the learned adjusted probabilities after running the linear program, $\mathbf{P^a}$ are guaranteed to be fair, taking the max value over the learned probabilities when predicting on an individual level will not maintain fairness. In fact, it can occur that taking the max over the adjusted probabilities will just result in identical predictions as those made by the original blackbox classifier. Instead, when predicting the class of an individual, the corresponding learned adjusted probabilities must be sampled from in order to maintain the fairness guarantee.



\section{Related Work}
Most prior work done on post-processing based fairness approaches focus on binary task prediction. \citet{wei2019optimized} create a post-processing algorithm that modifies the raw scores of a binary classifier (instead of thresholded hard predictions) in order to achieve desired fairness constraints expressed as linear combinations of the per-group expected raw scores. \citet{ye2020unbiased} develop a general in-processing fairness framework which alternates between a process of selecting a subset of the training data and fitting a classifier to that data. 

Several adversarial approaches to multiclass fairness have been investigated recently; although these are not blackbox post-processing algorithms. \citet{zhang2018mitigating} first present the idea of adversarial debiasing, while \citet{romano2020achieving} present a multiclass approach for in-process training based on adversarial learning, with the discriminator distinguishing between the distribution of the model's current predictions, the true label, and artificial protected attributes resampled to be fair, and the true distribution of the predictions, true labels, and true protected attributes.

Multiclass blackbox post-processing techniques are less studied; although there have been a few new approaches recently. Notably, \citet{denis2021fairness} derive an optimally fair classifier from a pre-trained model and show several nice theoretical guarantees, including the asymptotic fairness of their proposed plug-in estimator. We see 3 key differences between their approach and the extension to \citet{hardt2016equality} that we propose: they only consider binary protected attributes ($|\mathscr{A}|=2$), while we allow categorical protected attributes ($|\mathscr{A}| > 2$) and can take on any number of unique values, at least theoretically; their method requires fitting a new estimator to the test data, whereas ours only requires computing probabilities and solving a linear program, which is relatively efficient; and, perhaps most importantly, their approach is limited to the demographic parity fairness constraint, whereas our approach applies to any constraint that is linear in $\mathbf{P^a}$.


In broader terms, \citet{hossain2020designing} unify many of the published methods for learning fair classifiers by showing that equalized odds, equal opportunity, and other common measures of fairness in the binary setting are subsumed by their proposed generalizations of the economic notions of envy-freeness and equitability. They show that these generalizations of fairness apply to the multiclass setting, but post-processing techniques are incapable of achieving them. We show here that this notion is not entirely correct, at least in a narrow sense, and that fairness can be achieved with post-processing techniques in the multiclass setting, so long as the joint distribution $P(Y, \hat{Y}, A)$ is either fully known or can be reasonably approximated by a large-enough sample of training data.

\begin{table*}
    \footnotesize
    \centering
    \footnotesize
    \begin{tabular}{cccc}
     \multicolumn{4}{c}{\bf Experiments with $\mathbf{|\mathscr{A}|=3}$}\\
    \toprule \textbf{Hyperparameter} & \textbf{Level} & \textbf{Change in Acc (CI)} & \textbf{Change in $\mathbf{TDR}$ (CI)}
        \\
        \midrule
        Intercept & --  & -0.13 (-0.17, -0.09) & -0.18 (-0.21, -0.15)
        \\
        \midrule
        Loss & Unweighted & -- & -- 
        \\ & Weighted & -0.11 (-0.13, -0.09) & 0.12 (0.10, 0.13)
        \\
        \midrule
        Goal & Equalized Odds & -- & -- 
        \\ & Demographic Parity & 0.24 (0.22, 0.27) & 0.21 (0.18, 0.23)
        \\ & Equal Opportunity & 0.08 (0.05, 0.11) & 0.03 (0.01, 0.05)
        \\ & Term-by-Term & 0.08 (0.05, 0.11) & 0.02 (-0.01, 0.04)
        \\
        \midrule
        Group Balance & No Minority & -- & -- 
        \\ & One Slight Minority & -0.03 (-0.06, 0.00) & -0.02 (-0.04, 0.01)
        \\ & One Strong Minority & -0.04 (-0.07, -0.00) & -0.01 (-0.03, 0.02)
        \\ & Two Slight Minorities & -0.05 (-0.08, -0.02) & -0.02 (-0.04, 0.01)
        \\ & Two Strong Minorities & -0.07 (-0.11, -0.04) & -0.01 (-0.04, 0.01) 
        \\
        \midrule
        Class Balance & Balanced & -- & -- 
        \\ & One Rare & 0.02 (-0.00, 0.04) & -0.04 (-0.06, -0.02) 
        \\ & Two Rare & 0.07 (0.04, 0.09) & -0.18 (-0.20, -0.17)
        \\
        \midrule
        Pred Bias & Low One & -- & -- 
        \\  & Low Two & 0.00 (-0.03, 0.04) & -0.00 (-0.03, 0.02) 
        \\ & Medium One & -0.06 (-0.09, -0.02) & -0.06 (-0.08, -0.03) 
        \\ & Medium Two & -0.04 (-0.07, -0.00) & -0.06 (-0.08, -0.03) 
        \\ & High One & -0.18 (-0.22, -0.15) & -0.16 (-0.19, -0.14) 
        \\ & High Two & -0.15 (-0.19, -0.12) & -0.13 (-0.16, -0.11) 
        \\
        \midrule
    \end{tabular}
    \caption{Predicted change and 95\% confidence intervals for accuracy and mean $TDR$ as a function of the experimental hyperparameters in our synthetic datasets with three protected attributes. All datasets had a 3-class outcome.}
\end{table*}

\section{Synthetic Data Experiments}
\paragraph{Synthetic Data}
To explore the effect of different data regimes and optimization goals on post-adjustment discrimination, we conducted thorough (though by no means exhaustive) synthetic experiments for a 3-class outcome. We constructed synthetic datasets with $N=1,000$ observations for each unique combination of the following data-generating hyperparameters:
\begin{itemize}
    \item The number of unique values for the protected attribute, $|\mathscr{A}|$. We explored setting $|\mathscr{A}|=2$ or $|\mathscr{A}|=3$ (see results with $|\mathscr{A}|=2$ in our github repository)
    \item The amount of class imbalance for the labels $Y$. For simplicity, we did not allow this to vary across protected groups. 
    \item Group balance, or the number and relative size of minority groups compared to majority groups. This varied according to the number of groups but was generally either none, weak, or strong.
    \item Predictive bias as the difference in mean true detection rate, $TDR$, between the groups. We vary this from mild predictive bias (10 percent difference) to severe bias with the minority group $TDR$ being near chance. The predictive bias is set to always favor the majority group.
\end{itemize}

This process yielded 117 datasets. For each one, we ran the linear program to adjust the (synthetic) biased blackbox predictions 8 times, once for each unique combination of the objective function and type of fairness, yielding a total of 936 adjustments. After each adjustment, we recorded two broad measures of the fair predictor's performance:
\begin{itemize}
    \item Triviality, or whether any of the columns in $\mathbf{W^a}=Pr(Y^{\text{adj}}|Y, A=a)$ contained all zeroes (i.e., whether any levels of the outcome were no longer predicted). 
    \item Discrimination, or the percent change in loss for the adjusted predictor relative to that of the original predictor. For this measure, we examined two specific metrics: global accuracy and the mean of the group-wise $TDRs$. These are equivalent to 1 minus the post-adjustment loss under the two versions of the objective functions we present above.
\end{itemize}

To quantify the average effect of each hyperparameter on discrimination, we fit two multivariable linear regression models to the resulting dataset, one for each discrimination metric. Before fitting the models, we converted the categorical hyperparameters (so all but loss) to one-hot variables, and then we set a reference level for each, removing the corresponding column from the design matrix. We then fit the models separately using ordinary least squares (OLS) and calculated confidence intervals (CIs) for the resulting coefficients.


\paragraph{Results}
Table 1 shows coefficients and 95\% confidence intervals for the regression models with $|\mathscr{A}|=3$. The results highlight several important points:

\begin{itemize}
    \item Predictive bias and class imbalance are the two main drivers of decreases in post-adjustment discrimination, for both accuracy, and $TDR$. 
    \item High group imbalance for the protected attributes lowers post-adjustment discrimination, but only from the perspective of global accuracy--even with 2 strong minorities (3-group scenario), mean $TDR$ only drops by 1.1\%.
    \item Relative to the weighted objective, the unweighted objective leads to higher scores for global accuracy but lower scores for mean $TDR$. This is perhaps unsurprising, but it is worth noting nonetheless.
    \item Despite finding better accuracy solutions, we also found that the unweighted objective leads to trivial solutions far more frequently (30\% of the time it was used) than the weighted version of the loss (0.2\% of the time it was used). This trend will likely worsen with increasing dimension of either the number of classes or the number of protected groups. 
    \item Fairness is generally harder to achieve with 3 protected groups than with 2, since the intercepts are lower for both accuracy and mean $TDR$. We believe this to be a general consequence of forcing fairness across more groups and expect this trend to continue as the number of groups increases.

\end{itemize}

\begin{figure*}[!h]
    \centering
    \includegraphics[scale=.5]{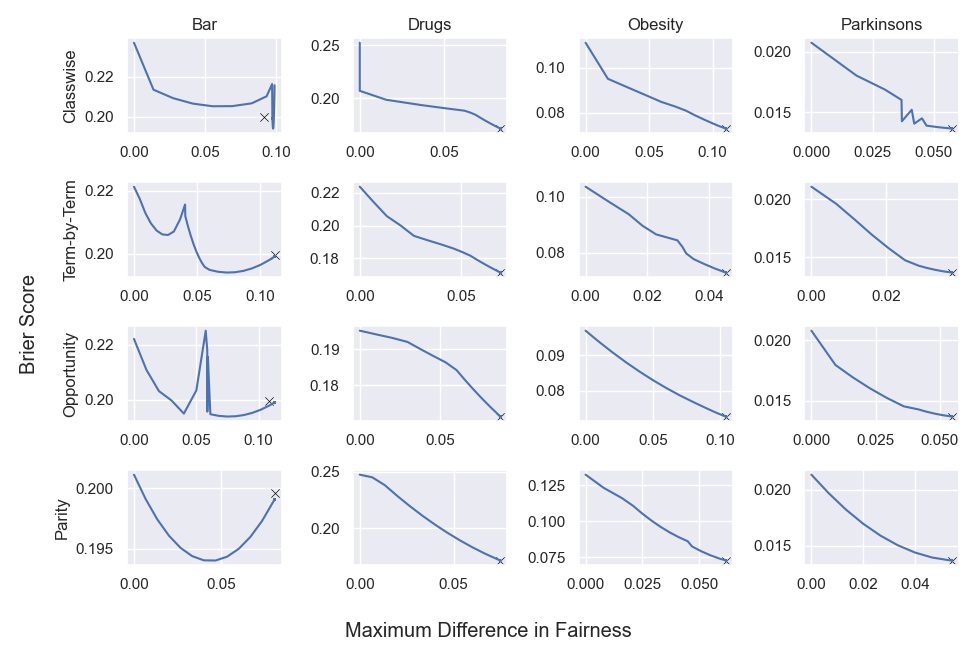}
    \caption{Fairness-discrimination plots for our postprocessing algorithm on our 4 real-world datasets, created by systematically relaxing the fairness equality constraints of the linear program. The plots show Brier score as a function of the maximum average difference between groups of the corresponding fairness criterion. Performance of the original, unadjusted predictor is marked by an X. 
    }
    \label{fig:fairness_vs_discrimination}
\end{figure*}

\section{Experiments with Real-World Data}
\paragraph{Dataset Descriptions}
To further examine the performance characteristics of our algorithm, we ran it on several real-world datasets described below. 
\begin{enumerate}
    \item \textbf{Drug Usage} \cite{fehrman2017five}. This dataset has inherently multiclass outcomes, with the target being a 7-level categorical variable indicating recentness of use for a variety of drugs. We focus on predicting cannabis usage, where we collapsed the 7-level usage indicator into 3 broader categories: never used, used but not in the past year, and used in the past year. Predictors included demographic variables like age, gender, and level of education, as well as a variety of measures of personality traits hypothesized to affect usage habits.
    \item \textbf{Obesity} \cite{palechor2019dataset}. This dataset has inherently multiclass outcomes, with the target being a 7-level categorical variable indicating weight category; the protected attribute is gender (Male/Female). Because some of the observations are synthetic in order to protect privacy, not all of the gender/weight categories had sufficient numbers for modeling, and so we omitted observations from the 2 most extreme weight categories, Obesity Type-II and Obesity Type-III, leaving a 5-level target for prediction. Predictors included age, gender, family medical history, and several measures of physical activity and behavioral health.
    \item \textbf{LSAC Bar Passage} \cite{wightman1998lsac}. This dataset has inherently multiclass outcomes, with the target being a 3-level variable indicating bar exam passage status (passed first time, passed second time, or did not pass). The protected attribute is race, which we collapsed from its original 8 levels to 2 (white and non-white). Predictors included mostly measures of educational achievement, like undergraduate GPA, law school GPA, and LSAT score.
    \item \textbf{Parkinson's Telemonitoring} \cite{tsanas2009accurate}. This dataset does not have inherently multiclass outcomes, with the target for prediction being the continuous Unified Parkinson's Disease Rating Scale (UPDRS), a continuous score that increases with the severity of impairment. We again used Otsu's method to bin the continuous score into 3 categories--low impairment, moderate impairment, and high impairment--which we took as the new class labels. The protected attribute is a 2-level variable for gender (Male/Female). Predictors included mostly biomedical measurements from the voice recordings of patients with Parkinson's Disease.
\end{enumerate}

For each of these datasets, we obtained a potentially-biased predictor $\hat{Y}$ by training a random forest on all available informative features (including the protected attribute) to predict the multiclass outcome, and then taking the categories corresponding to the row-wise maxima of the out-of-bag decision scores as the set of predicted labels. We then adjusted the predictions with the weighted objective and term-by-term equality of odds fairness constraint and recorded the relative changes in global accuracy and mean $TDR$ as the outcome measures of interest, as with our synthetic experiments.

\begin{table*}
    \footnotesize
        \centering
        \hspace{.61 cm}
        \begin{tabular}{c|cccc}
        \multicolumn{5}{c}{\bf In-Sample Results}\\
        \toprule
        Dataset (N) & \# Terms& Old Acc $\shortrightarrow$ New Acc & Old $TDR\shortrightarrow$ New $TDR$& Pre $\shortrightarrow$ Post-Adj Disparity\\
        &in $\mathbf{P^a}$&(\% change)&(\% change)&(\% change)\\
        \midrule
         Bar (N=22406)& 18&88 \% $\shortrightarrow$ 88\% (-1\%)& 36\% $\shortrightarrow$ 34\% (-7\%)&0.11 $\shortrightarrow$0.00 (-100\%)\\ 
         Cannabis (N=1885)&18&74\% $\shortrightarrow$ 71\% (-4\%)& 67\% $\shortrightarrow$ 63\% (-6\%)&0.07 $\shortrightarrow$0.00 (-100\%)\\ 
         Obesity (N=1490)&50&78\% $\shortrightarrow$ 73\% (-7\%)& 78\% $\shortrightarrow$ 73\% (-7\%)&0.05 $\shortrightarrow$ 0.00 (-100\%)\\ 
         Parkinsons (N=5875)&18&93\% $\shortrightarrow$ 91\% (-2\%)& 92\% $\shortrightarrow$ 89\% (-3\%)&0.04 $\shortrightarrow$0.00(-100\%)\\ 
        \end{tabular}
        \newline
        \vspace{.5 cm}
        \newline
        \begin{tabular}{c|cccc}
        \multicolumn{5}{c}{\bf Out of Sample Results}\\
        \toprule
        Dataset (N) & \# Terms& Old Acc $\shortrightarrow$ New Acc & Old $TDR\shortrightarrow$ New $TDR$& Pre $\shortrightarrow$ Post-Adj Disparity\\
        &in $\mathbf{P^a}$&(\% change)&(\% change)&(\% change)\\
        \midrule
         Bar (N=22406)& 18&88 \% $\shortrightarrow$ 83\% (-6\%)& 36\% $\shortrightarrow$ 33\% (-8\%)&0.11 $\shortrightarrow$0.01 (-95\%)\\ 
         Cannabis (N=1885)&18&74\% $\shortrightarrow$ 61\% (-18\%)& 67\% $\shortrightarrow$ 52\% (-22\%)&0.07 $\shortrightarrow$0.16 (124\%)\\ 
         Obesity (N=1490)&50&78\% $\shortrightarrow$ 41\% (-47\%)& 78\% $\shortrightarrow$ 42\% (-46\%)&0.05 $\shortrightarrow$ 0.07 (45\%)\\ 
         Parkinsons (N=5875)&18&93\% $\shortrightarrow$ 82\% (-12\%)& 92\% $\shortrightarrow$ 78\% (-15\%)&0.04 $\shortrightarrow$0.05(33\%)\\ 
        \end{tabular}
    \caption{Results of applying the linear program to adjust the blackbox predictions and produce $y^{der}$ for four real-world datasets.  The top table is without any splitting. Results shown in the bottom table are cross-validated across five 80/20 splits of each dataset. Accuracy and $TDR$ are shown before and adjustment, with $TDR$ being the mean across all classes. Percent changes, shown in parentheses are the relative percent drops in accuracy and mean $TDR$. Post-adjustment disparity is the element-wise mean difference across all groups of $\mathbf{W^a}$.}
\end{table*}

\paragraph{Exploring the Effect of Finite Sampling}
\citet{hardt2016equality} note that their method will not be effected by finite sample variability as long as the joint distribution $Pr(Y, \hat{Y}, A)$ is known, or at least well-approximated by a large sample. In practical applications, however, the sample at hand may not be large enough to approximate the joint distribution with precision. This problem is exacerbated when the number of observations $N$ is small relative to the number of probabilities learned by the algorithm of which there are $|C|\times|C|\times|\mathscr{A}|$ total. This difficulty is therefore more severe for our extension in this work where $|C| > 2$.



In these cases, the adjusted predictor $Y^{\text{adj}}$ may have worse classification performance and higher disparity when applied to unseen, out-of-sample data.
As a preliminary exploration of this effect, we used 5-fold cross-validation to generate out-of-sample predictions for each of the observations in our real-world datasets. Keeping $Y$, $\hat{Y}$, and $A$ fixed, we solved the linear program on $80\%$ of the data and then used the adjusted probabilities $\mathbf{P^a}$ to obtain class predictions for the observations in the remaining $20\%$. As with the predictions obtained from solving the linear program on the full dataset, we measured the changes in accuracy and mean $TDR$ for the cross-validated predictions. Because fairness is not guaranteed when the joint distribution assumption is violated, we also measured post-adjustment fairness.

\paragraph{Exploring the Fairness-Discrimination Tradeoff}
When there are large gaps in a predictor's performance across groups, i.e., when predictive bias is high, strict fairness may not always be possible or desirable to achieve because of the large amount of randomization required to balance the blackbox classifier's predictions. To explore the tradeoff between fairness and discrimination, we ran the linear program on each of the real-world datasets once for each of the four kinds of fairness. For each combination of dataset and fairness type, we varied the equality constraints of the linear program--the maximum percent difference allowed between any pairwise comparison of fairness measures between groups--from 0.0 to 1.0 in increments of 0.01, and then plotted the value of the weighted objective at each point as a function of the global measure of fairness corresponding to the fairness type under consideration. To obtain these global measures, we took the maximum of the mean differences across pairs of groups of the following metrics:

\begin{itemize}
    \item $\mathbf{W}$, or the matrix of probabilities $P(Y^{\text{adj}}|Y)$, for term-by-term equality of odds
    \item Youden's J index, or $TDR + (1-FDR) - 1$, for classwise equality of odds
    \item $TDR$ for equal opportunity
    \item $P(Y^{\text{adj}})$ for demographic parity
\end{itemize}

We note here that taking the maximum of the maxima of the pairwise differences would also be a valid and sensible global measure. So that the plots show performance under optimal conditions, we do not use cross-validation to obtain $Y^{\text{adj}}$, i.e., we obtain it by solving the linear program on the entire dataset.

\paragraph{Results}
Table 2 shows changes in global accuracy and mean $TDR$ after adjustment with the weighted objective and term-by-term conditional fairness constraint for our four datasets, using cross-validation as described above to capture some of the variability that comes with finite sampling. Overall, adjustment lowered both accuracy and mean $TDR$. Although, for the bar passage, drug usage, and Parkinson's datasets, the drops were moderate, with average relative changes in both metrics coming in at around 12\% and 15\%, respectively (without cross-validation, the drops were much smaller at 3\% and 4\%). For the obesity dataset, the drops are much larger at 47\% and 46\%, respectively, which are indeed substantial and would likely make the predictor unusable in practical settings. On in-sample data, these drops were both only around 7\%, and so we suspect that characteristics of the data, like large class imbalance or small overall sample size, are responsible for the poor performance.  Perhaps most importantly, the post-adjustment disparity for all datasets is non-zero, and for three of the datasets actually increases. The bar passage dataset was the only example where the out-of-sample post-adjustment disparity decreased to near zero likely due to it being the largest dataset. This starkly points out the sensitivity of the method to estimating the joint probabilities $Pr(Y, \hat{Y}, A)$, and shows that the approach is unlikely to work in smaller dataset regimes which have a larger combination of classes and protected attributes. Note that for in-sample results, post-adjustment disparity drops completely to 0.0 for all datasets since it is strictly enforced by the linear program in Table 2. 


Figure 1 shows fairness-discrimination plots for our 4 datasets with the weighted objective and each of the 4 fairness constraints. Under strict fairness, with inequality set to 0, equalized odds is the hardest to satisfy, showing the largest increase in Brier score. For the drug usage, obesity, and Parkinson's datasets, discrimination improves approximately linearly as fairness worsens; for the bar passage dataset, discrimination improves to a point, but then worsens as fairness approaches the value for the original, unadjusted predictor $\hat{Y}$. For all datasets, the total loss of discrimination under strict fairness is relatively small (the biggest drop is around 7.5 percentage points on Brier score), but the random forests' predictions were only mildly biased to begin with, so we expect this gap to increase for less-fair predictors.

\section{Discussion}
 Generally, our post-processing approach to achieving fairness in multiclass settings seems both feasible and efficient given a large enough dataset size. We have shown above that the linear programming technique proposed by \citet{hardt2016equality} can be extended to accommodate a theoretically arbitrarily large number of discrete outcomes and levels of a protected attribute. Nonetheless, our synthetic experiments and analyses of real-world datasets show that are a few important considerations for using the approach in practice. 
 
In many cases, the effect of finite sampling may be non-negligible, especially when the number of observations $N$ is small relative to number of outcomes $|C|$ or the number of protected groups $|\mathscr{A}|$. For example, the obesity dataset with $|C|=5$ and $N=1,490$ saw a large relative drop of 46\% in mean $TDR$ after adjustment under cross-validation.
We also saw this effect extend to fairness, which was not reduced completely to zero on out-of-sample data for any of the real-world datasets. In fact, for the drug usage dataset we found post-adjustment disparity doubled on out-of-sample data. 


This last observation raises a concerning point: for some classification problems, the post-adjustment predictions on out-of-sample data may increase disparity rather than lowering it. For the largest of the datasets, the bar passage dataset with $N=22,406$, neither of these issues was a concern. Even under cross-validation, the relative change in $TDR$ was only -8\%, and the disparity dropped to near 0 (-95\% decrease). Given this, we expect that with a large enough dataset size, our approach will be far more reliable on out-of-sample data. Future work more precisely quantifying the number of training examples needed for reliable out-of-sample fair performance with our approach is needed. 
 
More generally, even when finite sampling variability is not an issue, not all datasets will lend themselves well to this kind of post-processing approach. In our synthetic experiments, we showed that severe class imbalance and severe predictive bias (predicting at nearly the level of chance for minority protected groups) lead to large drops in post-adjustment performance on average. In many of the single experimental runs for synthetic datasets with these settings, the resulting derived predictor was effectively useless, either producing trivial results or lowering predictive performance to near chance (for all groups) for one or more class outcomes. In these circumstances, it may be more sensible to enforce fairness through a combination of pre-processing, in-processing, and post-processing methods, rather than through a post-processing method alone. Indeed, \citet{woodworth2017learning} make this point generally, albeit for the binary setting, by showing that unless the biased predictor $\hat{Y}$ is very close to being Bayes optimal, the derived predictor $Y^{\text{adj}}$ proposed by \citet{hardt2016equality} can underperform relative to other methods, sometimes substantially. Under less extreme circumstances, however, we found our approach produces good results, especially given the time-efficiency of solving the linear program relative to other methods.


\section{Acknowledgments}
This work was supported in part by the HPI Research Center in Machine Learning and Data Science at UC Irvine (P. Putzel), as well as in part by an appointment to the Research Participation Program at the Centers for Disease Control and Prevention, administered by the Oak Ridge Institute for Science and Education (P. Putzel). We would also like to thank Chad Heilig, and Padhraic Smyth for their helpful comments on the approach and paper.
\bibliography{citations.bib}

\begin{thebibliography}{17}
\providecommand{\natexlab}[1]{#1}

\bibitem[{Buolamwini and Gebru(2018)}]{buolamwini2018gender}
Buolamwini, J.; and Gebru, T. 2018.
\newblock Gender shades: Intersectional accuracy disparities in commercial
  gender classification.
\newblock In \emph{Conference on fairness, accountability and transparency},
  77--91. PMLR.

\bibitem[{Calders, Kamiran, and Pechenizkiy(2009)}]{calders2009building}
Calders, T.; Kamiran, F.; and Pechenizkiy, M. 2009.
\newblock Building classifiers with independency constraints.
\newblock In \emph{2009 IEEE International Conference on Data Mining
  Workshops}, 13--18. IEEE.

\bibitem[{Denis et~al.(2021)Denis, Elie, Hebiri, and Hu}]{denis2021fairness}
Denis, C.; Elie, R.; Hebiri, M.; and Hu, F. 2021.
\newblock Fairness guarantee in multi-class classification.
\newblock \emph{arXiv preprint arXiv:2109.13642}.

\bibitem[{Dieterich, Mendoza, and Brennan(2016)}]{dieterich2016compas}
Dieterich, W.; Mendoza, C.; and Brennan, T. 2016.
\newblock COMPAS risk scales: Demonstrating accuracy equity and predictive
  parity.
\newblock \emph{Northpointe Inc}.

\bibitem[{Dwork et~al.(2012)Dwork, Hardt, Pitassi, Reingold, and
  Zemel}]{dwork2012fairness}
Dwork, C.; Hardt, M.; Pitassi, T.; Reingold, O.; and Zemel, R. 2012.
\newblock Fairness through awareness.
\newblock In \emph{Proceedings of the 3rd innovations in theoretical computer
  science conference}, 214--226.

\bibitem[{Fehrman et~al.(2017)Fehrman, Muhammad, Mirkes, Egan, and
  Gorban}]{fehrman2017five}
Fehrman, E.; Muhammad, A.~K.; Mirkes, E.~M.; Egan, V.; and Gorban, A.~N. 2017.
\newblock The five factor model of personality and evaluation of drug
  consumption risk.
\newblock In \emph{Data science}, 231--242. Springer.

\bibitem[{Hardt, Price, and Srebro(2016)}]{hardt2016equality}
Hardt, M.; Price, E.; and Srebro, N. 2016.
\newblock Equality of opportunity in supervised learning.
\newblock \emph{Advances in neural information processing systems}, 29:
  3315--3323.

\bibitem[{Hossain, Mladenovic, and Shah(2020)}]{hossain2020designing}
Hossain, S.; Mladenovic, A.; and Shah, N. 2020.
\newblock Designing fairly fair classifiers via economic fairness notions.
\newblock In \emph{Proceedings of The Web Conference 2020}, 1559--1569.

\bibitem[{Hu and Chen(2018)}]{hu2018short}
Hu, L.; and Chen, Y. 2018.
\newblock A short-term intervention for long-term fairness in the labor market.
\newblock In \emph{Proceedings of the 2018 World Wide Web Conference},
  1389--1398.

\bibitem[{Palechor and de~la Hoz~Manotas(2019)}]{palechor2019dataset}
Palechor, F.~M.; and de~la Hoz~Manotas, A. 2019.
\newblock Dataset for estimation of obesity levels based on eating habits and
  physical condition in individuals from Colombia, Peru and Mexico.
\newblock \emph{Data in brief}, 25: 104344.

\bibitem[{Romano, Bates, and Cand{\`e}s(2020)}]{romano2020achieving}
Romano, Y.; Bates, S.; and Cand{\`e}s, E.~J. 2020.
\newblock Achieving Equalized Odds by Resampling Sensitive Attributes.
\newblock \emph{arXiv preprint arXiv:2006.04292}.

\bibitem[{Tsanas et~al.(2009)Tsanas, Little, McSharry, and
  Ramig}]{tsanas2009accurate}
Tsanas, A.; Little, M.; McSharry, P.; and Ramig, L. 2009.
\newblock Accurate telemonitoring of Parkinson’s disease progression by
  non-invasive speech tests.
\newblock \emph{Nature Precedings}, 1--1.

\bibitem[{Wei, Ramamurthy, and Calmon(2019)}]{wei2019optimized}
Wei, D.; Ramamurthy, K.~N.; and Calmon, F. d.~P. 2019.
\newblock Optimized score transformation for fair classification.
\newblock \emph{arXiv preprint arXiv:1906.00066}.

\bibitem[{Wightman(1998)}]{wightman1998lsac}
Wightman, L.~F. 1998.
\newblock LSAC National Longitudinal Bar Passage Study. LSAC Research Report
  Series.

\bibitem[{Woodworth et~al.(2017)Woodworth, Gunasekar, Ohannessian, and
  Srebro}]{woodworth2017learning}
Woodworth, B.; Gunasekar, S.; Ohannessian, M.~I.; and Srebro, N. 2017.
\newblock Learning non-discriminatory predictors.
\newblock In \emph{Conference on Learning Theory}, 1920--1953. PMLR.

\bibitem[{Ye and Xie(2020)}]{ye2020unbiased}
Ye, Q.; and Xie, W. 2020.
\newblock Unbiased Subdata Selection for Fair Classification: A Unified
  Framework and Scalable Algorithms.
\newblock \emph{arXiv preprint arXiv:2012.12356}.

\bibitem[{Zhang, Lemoine, and Mitchell(2018)}]{zhang2018mitigating}
Zhang, B.~H.; Lemoine, B.; and Mitchell, M. 2018.
\newblock Mitigating unwanted biases with adversarial learning.
\newblock In \emph{Proceedings of the 2018 AAAI/ACM Conference on AI, Ethics,
  and Society}, 335--340.

\end{thebibliography}

\section*{Appendix A}
\paragraph{Derivation of Linearity of Fairness Constraints:} In order to obtain linearity in the protected attribute conditional probability matrices $\mathbf{P^a}$ we must find an expression of the form $\mathbf{W^a} = \mathbf{P^a} \mathbf{M^a}$:
\begin{align*}
    W^a_{ij}= & Pr(Y^{\text{adj}}=i|Y=j, A=a)   \\
     =& \sum_k Pr(Y^{\text{adj}}=j, \hat{Y}=k|Y=j, A=a) \\
     =& \sum_k Pr(Y^{\text{adj}}=i|Y=j, A=a, \hat{Y}=k)\\
     &\hspace{1 cm}\times Pr(\hat{Y}=k|Y=j, A=a) \\
     =& \sum_k Pr(Y^{\text{adj}}=i|\hat{Y}=k, A=a)  \\
     &\hspace{1 cm}\times Pr(\hat{Y}=k|Y=j, A=a)\\
     =& \sum_k P^a_{ik} Z^a_{kj}
\end{align*}
where $Z^a_{kj} = Pr(\hat{Y}=k|Y=j,A=a)$, and can be estimated empirically using the original predictions of the blackbox classifier. Thus we have $\mathbf{W^a} = \mathbf{P^a}\mathbf{Z^a}$. Moving from the third to fourth line requires the conditional independence assumption $Y^{\text{adj}} \perp Y | A, \hat{Y}$. This assumption is violated in cases where the blackbox predictions are weak, for example completely random, and can intuitively be thought of as requiring that the initial blackbox classifier has reasonable discriminative performance. In other words, relevant information for predicting $Y$ is contained in $\hat{Y}$.

Multiclass equality of opportunity only requires enforcing equality on the diagonals of $\mathbf{W^a}$, and therefore is linear in $\mathbf{P^a}$ as well.

Enforcing the classwise version of multiclass equality of odds requires enforcing equality of opportunity, which is already shown to be linear above, and also enforcing the overall false detection rates to be equal across protected groups. So in order for classwise multiclass equality of odds to be linear, the false detection rates must be linear in $\mathbf{P^a}$, shown below:
\begin{align*}
    FDR^a_c = & Pr(Y^{\text{adj}}=c|Y\neq c, A=a)\\
    =&\frac{Pr(Y^{\text{adj}} = c, Y\neq c, A=a)}{Pr(Y\neq c, A=a)}\\
    =& \sum_j \sum_{c' \neq c} \frac{Pr(Y^{\text{adj}}=c, Y=c', \hat{Y}=j, A=a)}{Pr(Y\neq c, A=a)}\\
    =& \sum_j \sum_{c' \neq c} \frac{P^a_{cj}Z^a_{jc'}Pr(Y=c', A=a)}{Pr(Y\neq c, A=a)}\\
    =&\sum_j P^a_{cj} \sum_{c' \neq c} \frac{Z^a_{jc'}Pr(Y=c', A=a)}{Pr(Y\neq c, A=a)}\\
    =& \sum_j P^a_{cj}V^a_{jc}
\end{align*}
where $V^a_{jc}=\sum_{c' \neq c} \frac{Z^a_{jc'}Pr(Y=c', A=a)}{Pr(Y\neq c, A=a)}$. This allows us to write the protected attribute conditional false detection rates as $\mathbf{FDR^a} = diag(\mathbf{P^a}\mathbf{V^a})$. As before, $\mathbf{V^a}$ can be computed from the empirical estimates of $\mathbf{Z^a}$, and $Pr(Y=i, A=j)$.

For multiclass demographic parity we can write:
\begin{align*}
    D^a =& Pr(Y^{\text{adj}}|A=a)\\
    =&\frac{1}{Pr(A=a)}\sum_k Pr(Y^{\text{adj}}, A=a, \hat{Y}=k)\\
    =&\sum_k Pr(Y^{\text{adj}}|\hat{Y}=k, A=a)\frac{Pr(\hat{Y}=k, A=a)}{Pr(A=a)}\\
    =& \sum_k Pr(Y^{\text{adj}}|\hat{Y}=k, A=a)Pr(\hat{Y}=k|A=a)\\
    =& \mathbf{P^a} Pr(\hat{Y}|A=a)
\end{align*}
which is again linear in $\mathbf{P^a}$, and the conditional probability vector $Pr(\hat{Y}|A=a)$ can be computed emprically.

\paragraph{Synthetic Experiment Results with $|\mathscr{A}|=2$}
\begin{table*}[!hbtp]
\scriptsize
\center
\begin{tabular}{cccc}
    \multicolumn{4}{c}{\bf Experiments with $\mathbf{|\mathscr{A}|=2}$}\\
    \toprule \textbf{Hyperparameter} & \textbf{Level} & \textbf{Change in Acc (CI)} & \textbf{Change in $\mathbf{TDR}$ (CI)} 
         \\ 
         \midrule 
         Intercept & -- & -0.08 (-0.13, -0.03) & -0.14 (-0.18, -0.10)
         \\ 
         \midrule
         Loss & Unweighted & -- & -- 
         \\ & Weighted & -0.09 (-0.12, -0.06) & 0.10 (0.08, 0.13)
         \\
         \midrule
         Goal & Equalized Odds & -- & -- 
         \\ & Demographic Parity & 0.20 (0.15, 0.24) & 0.17 (0.14, 0.21) 
         \\ & Equal Opportunity & 0.02 (-0.02, 0.07) & 0.02 (-0.02, 0.05) 
         \\ & Strict & 0.021 (-0.02, 0.07) & 0.01 (-0.03, 0.04) 
         \\
         \midrule
         Group Balance & No Minority & -- & -- 
         \\ & Slight Minority & -0.05 (-0.09, -0.01) & 0.01 (-0.02, 0.04)
         \\ & Strong Minority & -0.07 (-0.11, -0.03) & 0.00 (-0.03, 0.04) 
          \\
          \midrule
          Class Balance & Balanced & -- & -- 
         \\ & One Rare & -0.005 (-0.04, 0.03) & -0.05 (-0.08, -0.01) 
         \\ & Two Rare & 0.08 (0.04, 0.11) & -0.14 (-0.17, -0.11) 
         \\
         \midrule
         Predictive Bias & Low & -- & -- 
         \\ & Medium & -0.06 (-0.10, -0.03) & -0.09 (-0.12, -0.05)
         \\ & High & -0.20 (-0.24, -0.16) & -0.18 (-0.22, -0.15) 
         \\
         \midrule
    \end{tabular}
    \caption{Regression coefficients and 95\% confidence intervals for accuracy and mean $TDR$ as a function of the experimental hyperparameters for the synthetic datasets with two protected attributes and three possible outcomes.}

    \end{table*}

\end{document}